\documentclass[runningheads]{llncs} 

\usepackage{graphicx} 
\usepackage{times}
\usepackage{xcolor}
\usepackage{soul}
\usepackage[utf8]{inputenc}
\usepackage{placeins}
\usepackage{subfig}
\usepackage[export]{adjustbox}
\usepackage{amssymb}
\usepackage{amstext}
\usepackage{mathdots}
\usepackage{stmaryrd}
\usepackage{mathtools} 
\usepackage{braket}
\usepackage{url}
\usepackage[ruled]{algorithm2e}
\usepackage{lmodern}
\usepackage[T1]{fontenc}
\usepackage{textcomp}
\usepackage{tikz}
\usetikzlibrary{matrix}
\usepackage{float}
\usepackage[misc]{ifsym}

\DeclareMathOperator*{\sensors}{S}

\usepackage{comment}

\begin{document}


\title{
The Ecosystem Path to AGI\thanks{The computations were enabled by resources provided by the Swedish National Infrastructure for Computing (SNIC), partially funded by the Swedish Research Council through grant agreement no. 2018-05973.}
}

\author{
Claes Stranneg\aa rd\thanks{Corresponding author: \email{claes.strannegard@chalmers.se}} \and
Niklas Engsner \and
Pietro Ferrari \and
Hans Glimmerfors \and
Marcus Hilding Södergren \and
 Tobias Karlsson \and
 Birger Kleve \and
 Victor Skoglund 
}
\authorrunning{C. Stranneg\aa rd et al.}
%

\institute{Department of Computer Science and Engineering\\
Chalmers University of Technology and
University of Gothenburg 
}
\maketitle
\begin{abstract}
We start by discussing the link between ecosystem simulators and artificial general intelligence (AGI).
Then we present the open-source ecosystem simulator Ecotwin, which is based on the game engine Unity and operates on ecosystems containing inanimate objects like mountains and lakes, as well as organisms, such as animals and plants. Animal cognition is modeled by integrating three separate networks: (i) a \textit{reflex network} for hard-wired  reflexes; (ii) a \textit{happiness network} that maps sensory data such as oxygen, water, energy, and smells, to a scalar happiness value; and (iii) a \textit{policy network} for selecting actions. The policy network is trained with reinforcement learning (RL), where the reward signal is defined as the happiness difference from one time step to the next. All organisms are capable of either sexual or asexual reproduction, and they die if they run out of critical resources. 
We report results from three studies with Ecotwin, in which  natural phenomena emerge in the models without being hardwired. 
First, we study a terrestrial ecosystem with wolves, deer, and grass, in which a Lotka-Volterra style population dynamics emerges. 
Second, we study a marine ecosystem with phytoplankton, copepods, and krill, in which a 
diel vertical migration behavior emerges.
Third, we study an ecosystem involving lethal dangers, in which certain agents that combine RL with reflexes outperform pure RL agents.

\end{abstract}

\subsection*{Animal cognition}
All organisms in nature are subject to natural selection and, where applicable, also sexual selection \cite{darwin1859origin}. These forces operate on physical as well as cognitive properties of the organisms. 
One factor that contributes to the selection pressure on animal intelligence is other organisms. In fact, organisms coevolve with each other and are part of each other's environment. Intelligence may be an advantage in the “arms race'' between predators and prey, in the competition for scarce resources, in collaborations with mutual benefits, and in sexual selection processes. 
Another factor that contributes to the selection pressure on intelligence is the terrain. In fact, animals must continuously handle challenges imposed by the terrain and adapt their behavior to local conditions when migrating, foraging, approaching prey, escaping predators, mating, and parenting.

To survive and reproduce, animals need to deal with a continuous stream of challenges in their lives. They must find food, avoid predators, navigate, and mate. Solving these challenges requires efficient information processing, in particular perception, decision-making, and action. 
Nervous systems are present in almost all taxa of the animal kingdom and play a key role in animal information processing. They are typically far from monolithic. For example, in humans, the complex nervous system that controls digestion is essentially separated from the brain and may operate even in a brain-dead person. The human brain itself is highly modular, with its anatomically distinct lobes and regions, such as the \textit{Brain stem} which controls reflexes ---positive, e.g., the knee reflex, and negative, e.g., the diving reflex; the \textit{Prefrontal cortex}, which maps sensory signals to actions; and the \textit{Insula}, which combines internal signals, such as blood sugar level and external signals, such as smells, into signals linked to happiness and reward \cite{Insula2012}. 

 Reflexes are critical to many animals \cite{stein1988modulation}. For example, a newborn lamb might not have the time to find its way to its mother's udder through random trial and error. Instead, its early life hinges on an instinct that draws it to the smell of milk and a positive suck reflex that causes it to drink. The lamb might also benefit critically from negative reflexes that prevent it from eating lethal objects, inhaling liquid, or jumping from high cliffs.
From an evolutionary perspective, one of the great advantages of nervous systems is that they enable learning, i.e., physical modification, and thereby efficient adaptation to the dangers and resources of the local environment.  
A prominent example of learning is reinforcement learning (RL), which is used across the animal kingdom   \cite{niv2009reinforcement,neftci2019reinforcement}. 
%
Many animals combine untrainable reflex circuits with circuits that are trainable with RL. For instance, humans have hundreds of hardwired reflexes, e.g. the knee reflex, but we can also learn via RL. This enables us to combine the benefits of reflexes with the benefits of RL.

\subsection*{Models of animal cognition}
There is a great number of computational models of animal cognition. 
In the “reflex tradition'', animals are modeled as reflex agents, without the ability to learn. Such models might be adequate for animals with hardwired nervous systems that remain essentially unchanged during their lifetime. An example could be the famous nematode \textit{C. elegans} \cite{xu2020connectome}. Many biologically inspired algorithms, such as cellular automata, swarm algorithms, and ant algorithms, belong to this tradition and so does the subfield that studies populations of reflex agents powered by evolutionary algorithms. 

In the “RL tradition'', animals are modeled as RL agents. This field includes physical animal robots that get rewarded for fast crawling, running, swimming, or flying. 
It also includes homeostatic agents that get rewarded for keeping a set of homeostatic variables ---like energy, water, and oxygen--- close to their target values, or sweet spots \cite{keramati2011reinforcement,yoshida2017homeostatic}. 
RL algorithms have been shown to have a great potential for AI. For instance, they outperform humans at several video games and strategic board games \cite{badia2020agent57}. 
There are also several agent-based ecosystem simulators that are based on RL  
\cite{lanham2018learn,sunehag2019reinforcement,yamada2020evolution}.

In analytic approaches to ecosystem modeling, organisms are typically modeled with numbers representing population size or biomass and the interaction dynamics is modeled with systems of differential equations. There is no model of individual animals, no model of animal cognition, and no model of the terrain. Examples include the well-known Lotka-Volterra predator-prey dynamics \cite{lotka1925elements}. 
The Ecopath (with Ecosim and Ecospace) 
simulator for marine ecosystems  \cite{christensen2004ecopath} 
divides maps of ecosystem into geographical cells, where each cell contains populations, for example, given as tonnes of phytoplankton, zooplankton, planktivores, and pescivores. 


\subsection*{AI via models of animal cognition}
Nervous systems provided the original inspiration for the neural network model and its applications to supervised learning \cite{zador2019critique} and RL \cite{neftci2019reinforcement}.
Some researchers try to copy animal brains in wetware or reproduce their connectome in software \cite{xu2020connectome}, 
 others build computational models of the brain, sometimes called cognitive architectures \cite{kotseruba202040}.
%
Rather than aiming directly for a model of the brain, one may aim for a model of the process that led to the development of the brain. 
Since natural general intelligence emerged as the result of an evolutionary process that took place in an ecosystem, a natural strategy for creating AGI is to construct an ecosystem simulator that exploits the natural selection pressure on intelligence.
This strategy is in line with Wilson's ``animat path to AI'' \cite{wilson1991animat} and it enables a gradual approach to AI that starts with relatively simple ecosystems.

Section \ref{section:FW} of this paper presents the theoretical framework of the ecosystem simulator Ecotwin, which combines RL and reflexes. Section \ref{section:Results} presents simulation results, in which well-known patterns from population dynamics and ethology emerge.
Section \ref{section:Conclusion}, finally, draws some conclusions.

\section{Ecosystem simulator}
\label{section:FW}
In this section, we present the theoretical framework of the ecosystem simulator Ecotwin.
Ecotwin uses the game engine Unity with ML-Agents \cite{lanham2018learn}, which provides a graphical user interface, a physics engine, and several RL algorithms.
More details about Ecotwin, including its open source code, can be found at \url{www.ecotwin.se}. 

\subsection*{Ecosystems}
We model \textit{space} with a set $\mathcal{S} \subseteq \mathbb{R}^3$ and \textit{time} with the numbers $0, 1, 2, \dots$. 
The building blocks of our ecosystems are called \textit{objects}. For instance, we could introduce cat objects, dandelion objects, and rock objects. Each object is assigned a set of object properties.
\begin{definition}[Object properties] The object properties are:

\begin{itemize}
\item \emph{Physical properties} (measured in their respective SI units) such as 
Temperature,
Mass,
Pressure,
Electric current, and
Luminous intensity. Moreover, each object has a \textit{conformation}, which is a subset of $\mathcal{S}$. 
\item \emph{Chemical properties} (measured in the SI unit molarity) are concentrations of chemical substances such as 
Oxygen,
Nitrogen,
Water,
Carbon dioxide,
Glucose,
Salt,
Sulfur,
Fat,
Protein, 
Estrogen,
Testosterone, and
Oxytocin. 
\item \emph{Biological properties} such as
Age (seconds),
Sex (none, female, or male), and
Fertility (a real number in $[0,1]$).
\end{itemize}
\end{definition}
Note that the conformation of an object describes its shape and position in $\mathcal{S}$. The physical and chemical properties of objects may be defined as aggregations of the properties of the points inside their conformations, e.g., the average temperature of a rock object. It is sometimes convenient to introduce additional properties. For example, to talk about cat objects, one may want to introduce properties such as Eye color, Paw size, Blood pressure, and Blood sugar. 
\begin{definition}[Inanimate objects]
An \textit{inanimate object} in $\mathcal{S}$ consists of 
\begin{itemize}
\item A type: for example Rock, Water, Road, Building
\item A set of physical and chemical properties
\end{itemize}
\end{definition}

\begin{definition}[Organisms]
An \textit{organism} in $\mathcal{S}$ consists of
\begin{itemize}
\item A type: for example a particular species of bacteria, fungus, plant, or animal
\item A \textit{genome} consisting of a string over a finite alphabet
\item A set of physical, chemical, and biological properties

\item A \textit{nervous system} consisting of a set of sensors $\sensors$, where each sensor has a location in the conformation and a sensitivity to some physical or chemical property; a set of actions $A$, and three disjoint networks: a \textit{reflex network} with input nodes $\sensors$ and output nodes $A$; a \textit{policy network} with input nodes $\sensors$ and output nodes $A$; and a \textit{happiness network} with input nodes $\sensors$ and a single output node for representing a scalar happiness value. 
\item A set of \textit{hyperparameters}. These may include hyperparameters for training the policy network and for regulating reproduction and death. An example of a hyperparameter is $age\_{max}$, which controls the maximum lifetime.
\item A set of update rules:
\begin{itemize}
\item Update rules for physical properties, e.g. how locomotion actions influence the conformation. 
\item Update rules for chemical properties, e.g. how ingestion and locomotion actions influence the Glucose property. 
\item Update rules for biological properties specifying (i) the reproduction process, which may be sexual or asexual and involve mutation and crossover; and (ii) the physical and chemical properties that must be met for the organism to stay alive. For example, if Temperature is not inside a certain interval, or if $Age > age\_{max}$, 
then the organism will die and thus become an inanimate object. 
\end{itemize}
\end{itemize}
\end{definition}
The genome has two roles: 
(1) It encodes the organism at the start of its ``life'', in particular, it could encode its initial conformation (body shape), nervous system, and hyperparameters;
(2) It is the input to the reproduction process, in which one or two genomes give rise to a new genome. 

This definition of organism encompasses pure reflex agents (with an empty policy network), pure RL-agents (with an empty reflex network), and agents that combine reflexes with RL. 
By leaving the nervous system empty, we can also model agents lacking nervous systems altogether, such as bacteria and plants. 
It is often convenient to represent the reflex and happiness networks as hand-coded functions. If desired, these functions can easily be converted into neural networks, by using supervised learning. This step might be useful when applying mutations that add noise to connection weights.

\begin{definition}[Ecosystem]
An \textit{ecosystem} consists of
\begin{itemize}
\item A space $\mathcal{S}\subseteq \mathbb{R}^3$
\item A set of inanimate objects in $\mathcal{S}$
\item A set of organisms in $\mathcal{S}$.
\end{itemize}
\end{definition}
Several examples of ecosystems will be given in Section \ref{section:Results}. The ecosystem is updated at each tick, using the physics mechanisms of Unity and the update rules of the organisms. 

\subsection*{Decision-making} 
\begin{figure}
 \begin{center}
   \includegraphics[width=0.6\textwidth]{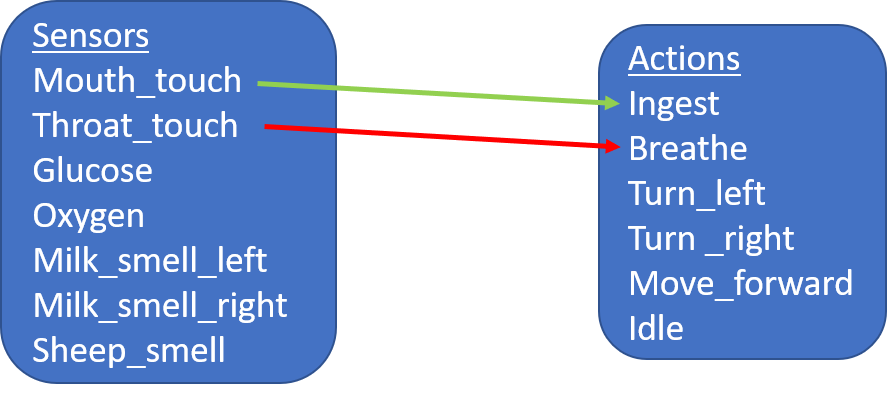}
 \end{center}
 \caption{A Reflex network. This network has sensors for touch (pressure) at two different locations of the conformation. It also has sensors for glucose and oxygen concentration in the blood stream, and for smell at different locations.
The actions are for homeostatic regulation and locomotion.
Only two connections have non-zero weights: one with weight +1 (green) and one with weight -1 (red). This produces a positive sucking reflex and a negative reflex that is similar to a diving reflex and prevents inhalation of liquid.
} 
\label{fig:nervous_system}
\end{figure}

This is how the nervous system is used for making decisions at time $t$: 
\begin{enumerate}
\item For each sensor, read off its physical or chemical property at its current position. This produces a vector $\textbf{x}$ that encodes the sensory input at $t$.
\item Give the input $\textbf{x}$ to the Reflex network, which then outputs some vector $\textbf{y}$, with component values -1, 0, or 1. The intended interpretation of these values are, respectively, to block, accept, and force the corresponding action. 
\item Also give the input $\textbf{x}$ to the Policy network, which then selects an action $a$. Represent this action as a one-hot vector $\textbf{z}$, where 1 is in the position of $a$. 
\item Finally, compute the vector $H(\textbf{y}+\textbf{z})$. Here, $H((v_1,\dots,v_n))$ is defined as $(h(v_1),\dots,h(v_n))$, where $h(v_i)$ is 1 if $v_i>0$ and 0 otherwise. This produces a multi-hot vector that constitutes the decision at $t$. Thus, the decision might be no action, one action, or many actions.  
\end{enumerate}
Now that we have seen how nervous systems are used for decision-making, let us consider a concrete example of a  nervous system of a lamb model. The sensors, actions, and reflex network are shown in Fig. \ref{fig:nervous_system}.
The happiness network maps Glucose and Oxygen to a happiness value, so that ingestion and breathing are encouraged. An additional input could be Sheep\_smell, so that the animal is encouraged to approach sheep: a hardwired social instinct.
The policy network is trained using RL and updated at each time step. Thus, the animal might learn to breathe when Oxygen is low and move toward the strongest smell of milk when Glucose is low.

\subsection*{Learning}
The reflex and happiness networks develop through evolution only, and not through learning. In contrast, the policy network is updated at each time step via RL. The reward signal at time $t$ is defined as $happiness(t)-happiness(t-1)$, where $happiness$ is computed by the happiness network. In the simulations discussed in this paper, we use the standard RL algorithm PPO \cite{schulman2017proximal} together with the animals' individual hyperparameters.

\section{Results}
\label{section:Results}

We will present several results that were obtained using Ecotwin simulations. Videos of the simulations are available at \url{www.ecotwin.se}.
%
Ecotwin can run on ecosystems populated with models of real organisms, imaginary organisms, or robots. 
In this section, we will consider simple models of real organisms. The properties and the mechanisms of our model organisms will be taken from the corresponding real organisms.
In order to find a reasonable starting point, we initialize the policy networks randomly and then pre-train them with RL in Ecotwin, before turning on reproduction and starting the simulations. 

\subsection*{Emergence of predator-prey dynamics}
\label{subsection:ecosys_1}

Previous work in predator-prey dynamics has shown that a two-species predator-prey system, with agents trained through RL, exhibited Lotka-Volterra cycles under certain choices of parameters \cite{yamada2020evolution}. We wanted to explore whether a three-species system would also exhibit such cycles. 

\begin{figure}[ht!]
    \includegraphics[width=0.47\textwidth, valign=t]{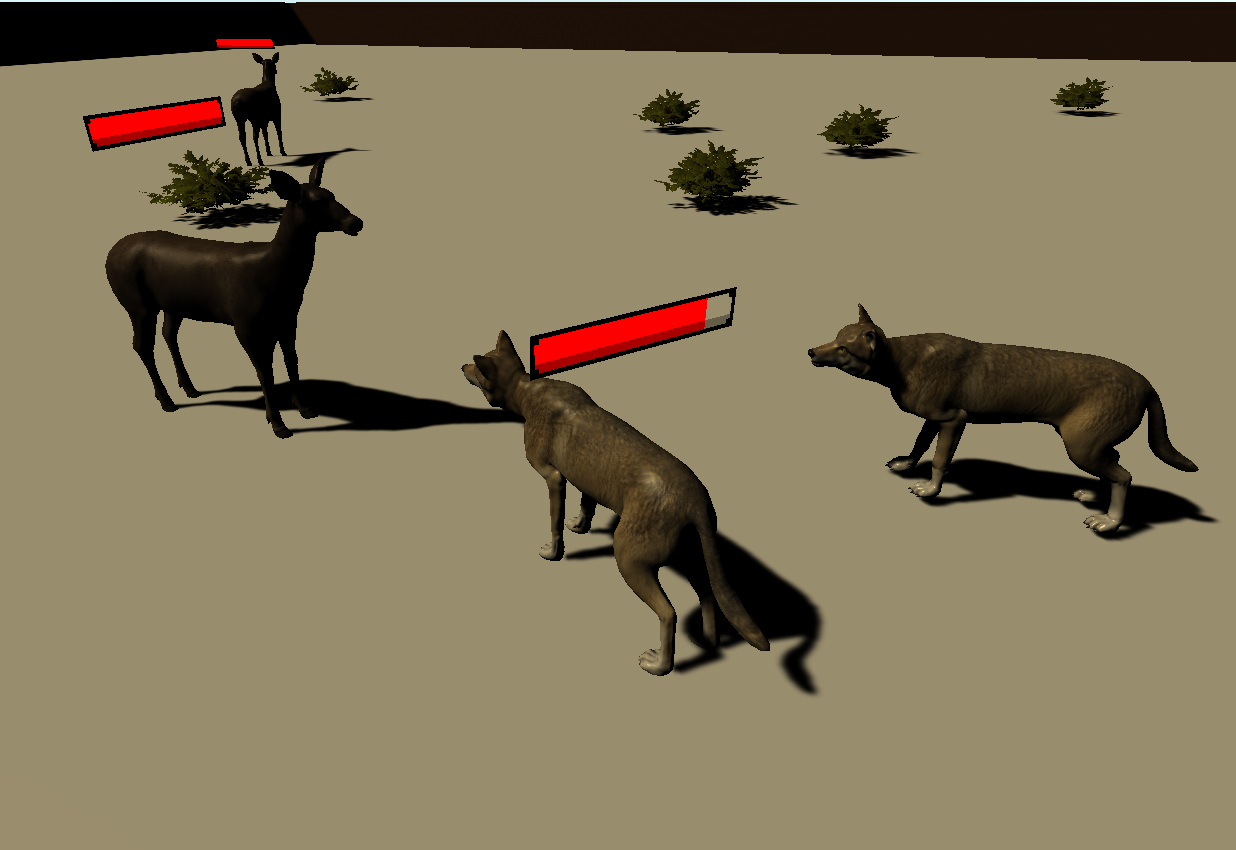}%
    \hfill
    \includegraphics[width=0.47\textwidth, valign=t]{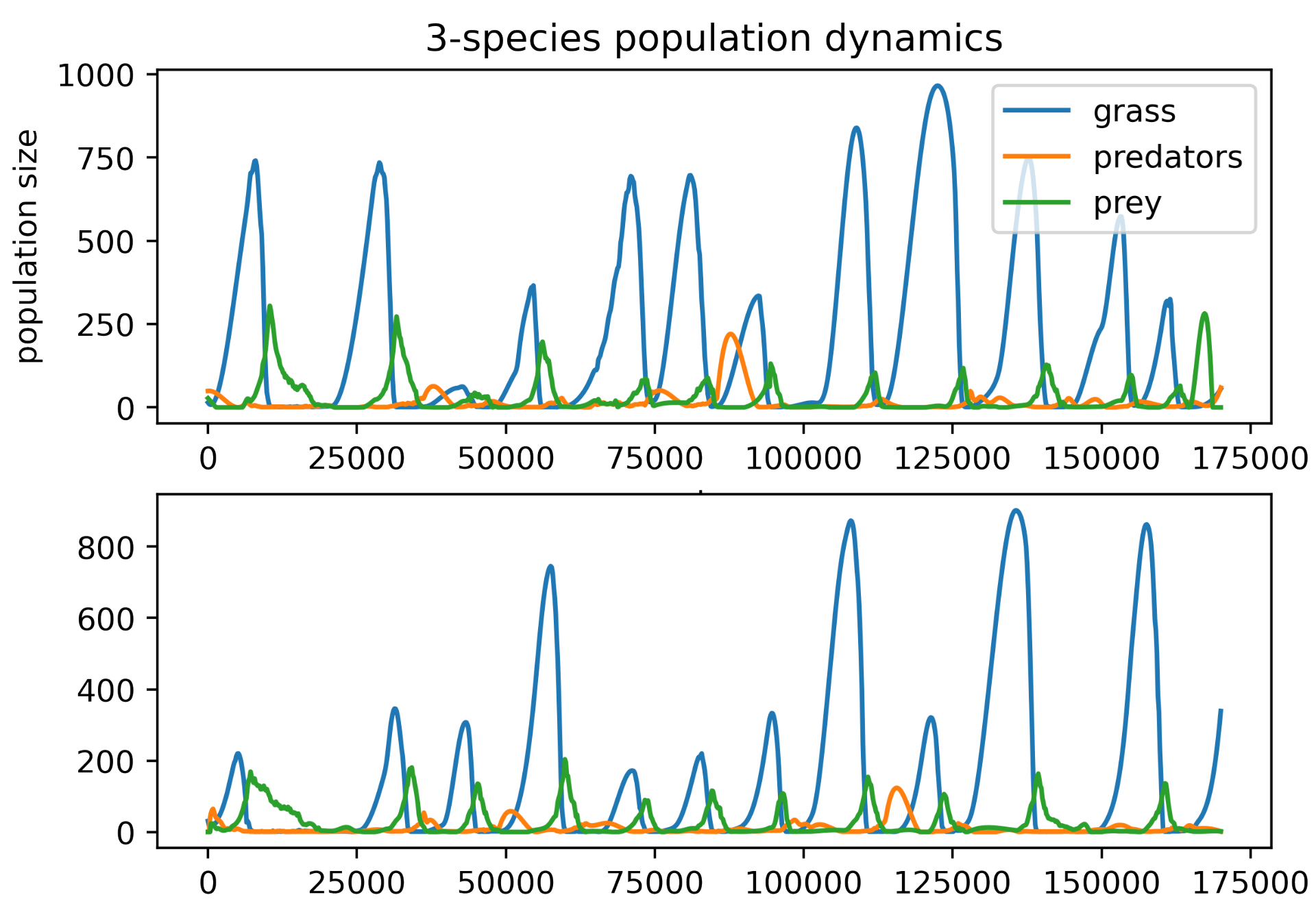}%
    \caption{
    \textbf{Left}: A three-species predator-prey ecosystem, with grass, deer and wolves. The red bars show happiness values.
    \textbf{Right}: Population dynamics from two different simulations. 
    } 
    \label{fig:lv}
\end{figure}

Our study concerns a three species predator-prey ecosystem with grass, deer, and wolves, as illustrated in Fig. \ref{fig:lv} (Left). 
Deer and wolves have vision, which is modeled via Unity's ray casts, and gives the direction and distance to the closest visible objects of each type, within a certain radius \cite{lanham2018learn}). Moreover, the deer can smell grass and wolves, while the wolves can smell deer.
 At each time step, each animal decides whether it should stand still, walk, or run forward. It also decides whether it should rotate left, right, or not at all. 
 The happiness of the deer is determined by their Energy and Wolf smell sensors, whereas happiness of the wolves is determined by their Energy and Deer smell sensors. 
The deer obtains energy by eating grass and wolves by eating deer. 
The consumption of energy depends on the velocity of the animal.
A simplified reproduction mechanism was used, where each animal had a probability of giving birth to a new agent, which was then spawned at a random unoccupied location in the ecosystem. 

We ran two simulations with Ecotwin, starting with the same ecosystem. The result is shown in Fig. \ref{fig:lv} (Right). As expected, given the dependence on randomness, the two simulations are different. In both simulations we see Lotka-Volterra cycles, with an increase in grass, followed by an increase in deer, followed by an increase in wolves, and similar decreases.
More details about the study can be found in \cite{karlsson2021}. 

\subsection*{Emergence of diel vertical migration}
\label{subsection:ecosys_3}

\begin{figure}[h!]
    \includegraphics[width=0.35\textwidth, valign=t]{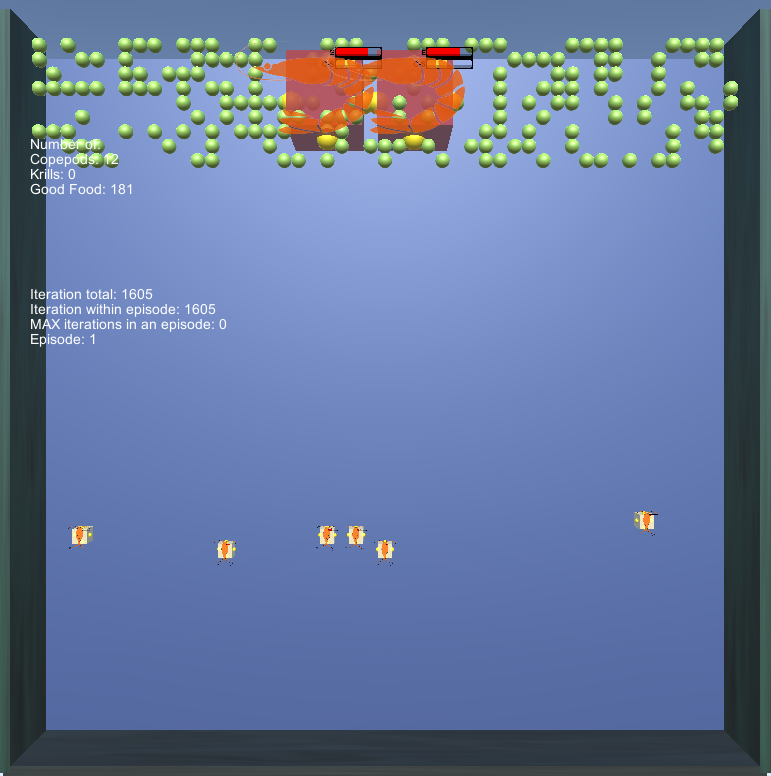}%
    \hfill
    \includegraphics[width=0.55\textwidth, valign=t]{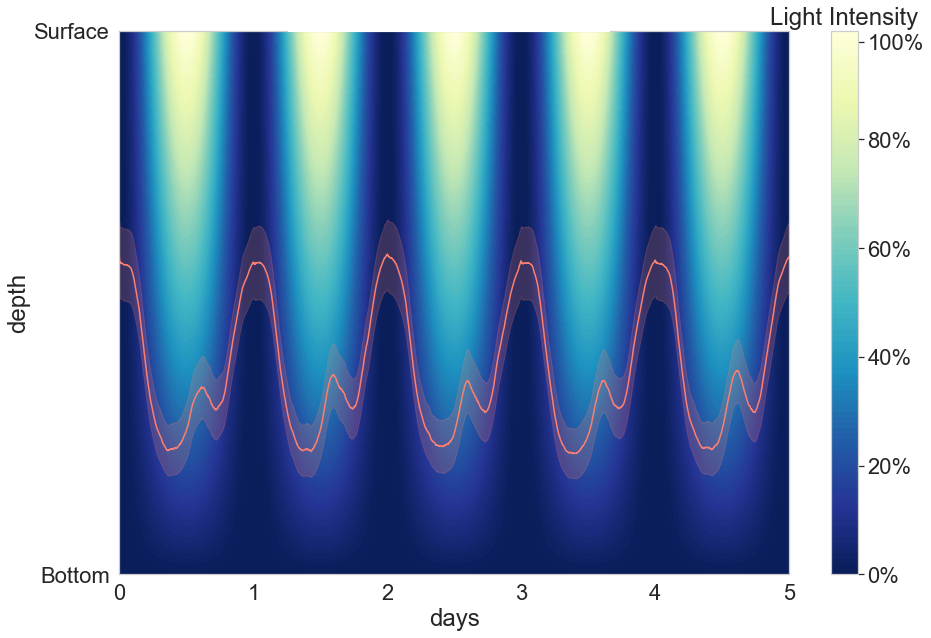}%
    \caption{
    \textbf{Left}: A marine environment with krill (orange), Copepods (yellow), and phytoplankton (green). The light intensity at different depths varies over time. This snapshot was taken at simulated daytime, when the Copepods are relatively far from the surface.
    \textbf{Right}: Mean vertical position of the copepods (red curve) over time. The light intensity is indicated by the background color. 
    } \label{fig:copepods}
\end{figure}

In this study, we consider a marine ecosystem with krill, copepods and phytoplankton, illustrated in Fig. \ref{fig:copepods} (Left). A diel light cycle was simulated, with the sun going up and down. We also simulated decreasing light intensity at greater depths.
Three behaviors observed in real Copepods \cite{seuront2014copepods} were studied:
(1) Diel Vertical Migration (DVM): a cyclic behavior in which Copepods migrate to the surface at night (enabling them to graze phytoplankton that are near the surface where the light is) and go down to greater and darker depths in the daytime (where the vision of predators such as krill is less efficient).  
(2) Quick escape reactions that enable Copepods to escape from predators. 
(3) Chemotaxis causing Copepods to avoid the smell of predators. 
 
The Copepod models have a simple form of vision, by which they can perceive a rough direction and distance to near-by phytoplankton. Moreover, they can perceive light intensity, their own energy level, and whether they are touching food, krill, the environment's boundaries,
or another Copepod. 
Their happiness value, yielding their reward, depends on energy and light intensity.

The Copepods and krill were pre-trained with RL for 1 million time steps. Then an Ecotwin simulation was run, starting with the ecosystem shown in Fig. \ref{fig:copepods} (Left). The behavioral patterns (1)-(3) were observed in the simulations. In particular, a clear DVM pattern was observed, as shown in Fig. \ref{fig:copepods} (Right). 
More information about the study can be found in \cite{kleve2021}.

\subsection*{Emergence of critical reflexes}
\label{subsection:ecosys_2}

\begin{figure}[ht!]
    \includegraphics[width=0.47\textwidth, valign=t]{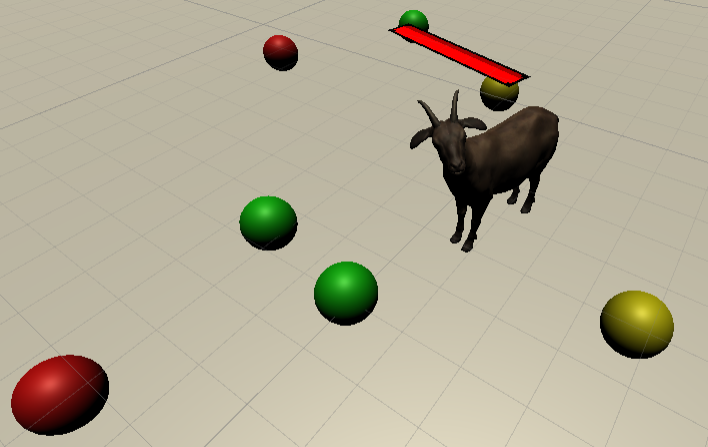}
    \hfill
   \includegraphics[width=0.47\textwidth, valign=t]{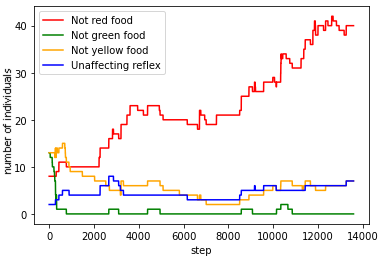}
    \caption{
    \textbf{Left}: An ecosystem with goats, and three different types of grass objects: green = good, yellow = bad but non-lethal, and red = lethal.
    \textbf{Right}: The number of goats that have the red, yellow, green, and blue genes.} \label{fig:reflexes}
\end{figure}

This study concerns the interplay between RL and reflexes. We consider a terrestrial ecosystem with goats and grass, illustrated in Fig. \ref{fig:reflexes} (Left). 
There are three types of grass in the environment: green, yellow and red. The green grass is good for the goats to eat, the yellow grass is bad, but not deadly, whereas the red grass is deadly. 
The goats reproduce sexually and have a genome that encodes their policy networks and reflex networks. Thus, they may pass on genomes, including genes that code for reflexes, to their offspring. The goats have four different genes that we call red, yellow, green, and blue, for convenience. These genes correspond to a reflex which prevents a goat from eating red, yellow and green food objects, respectively. The blue gene does not affect a goat's reflexes. 
  
An Ecotwin simulation was run with these organisms and the result is shown in Fig. \ref{fig:reflexes} (Right). We can see several patterns: 
(1) The red gene eventually dominates the population. This is expected, as a reflex to avoid lethal food gives them a clear advantage.
(2) The domination of the red gene suggests that the combination of reflexes and RL is more effective than RL alone when there are lethal dangers in the environment. Avoiding lethal food cannot be learned during a goat's life.
(3) The yellow gene does not have a clear advantage over the blue gene. This suggests that the combination of reflexes and RL gives no advantage, that cannot be learned, over pure RL-agents when the dangers are not lethal.
(4) The goats with the green gene keep dying out and then reappearing because of the mutation in the inheritance mechanisms.
(5) A goat's genome may include multiple genes. This explains why the superior red gene does not dominate the population completely.
More details about the study can be found in \cite{skoglund2021}. 

\section{Conclusion} 
\label{section:Conclusion}
We have presented the open source ecosystem simulator Ecotwin. 
It was run on three ecosystems, on which it reproduced certain population dynamics and behavioral patterns that can be observed in real ecosystems. 
Agent-based ecosystem simulators can be used for predicting the consequences of human interventions, e.g., via fishing, forestry, or urbanization. 
They might also be used as ``general AI gyms'', in which populations of agents coevolve in a fully automatic process, while taking advantage of the built-in selection pressure on intelligence. This research path toward general AI seems to be worthy of further exploration.

\bibliographystyle{ieeetr} 
\bibliography{animatreferences}

\end{document}